\documentclass[lettersize,journal]{IEEEtran}
\usepackage{amsmath, amsfonts}
\usepackage{amsthm}
\usepackage{algorithmic}
\usepackage{array}
\usepackage[caption=false,font=normalsize,labelfont=sf,textfont=sf]{subfig}
\usepackage{textcomp}
\usepackage{stfloats}
\usepackage{url}
\usepackage{verbatim}
\usepackage{graphicx}
\usepackage{hyperref}
\usepackage{float}
\usepackage{booktabs} 
\usepackage{diagbox}
\usepackage{multirow}
\usepackage{multicol}
\usepackage{wrapfig}
\usepackage{color}
\usepackage{svg}
\allowdisplaybreaks[4]

\newcommand{\red}[1]{\textcolor{red}{#1}}

\allowdisplaybreaks

\newtheorem{theorem}{Theorem}

\newtheorem{assumption}{Assumption}
\def\BibTeX{{\rm B\kern-.05em{\sc i\kern-.025em b}\kern-.08em
    T\kern-.1667em\lower.7ex\hbox{E}\kern-.125emX}}
\usepackage{balance}

\usepackage{cleveref}
\Crefname{theorem}{Theorem}{Theorems}

\begin{document}

\title{\textbf{Co}operative \textbf{R}eward \textbf{S}haping for Multi-Agent Pathfinding}
\author{
        Zhenyu Song, 
        Ronghao Zheng,
        Senlin Zhang, 
        Meiqin Liu
}


\maketitle
\pagestyle{empty}
\thispagestyle{empty}

\begin{abstract}
The primary objective of Multi-Agent Pathfinding (MAPF) is to plan efficient and conflict-free paths for all agents. Traditional multi-agent path planning algorithms struggle to achieve efficient distributed path planning for multiple agents. In contrast, Multi-Agent Reinforcement Learning (MARL) has been demonstrated as an effective approach to achieve this objective. By modeling the MAPF problem as a MARL problem, agents can achieve efficient path planning and collision avoidance through distributed strategies under partial observation. However, MARL strategies often lack cooperation among agents due to the absence of global information, which subsequently leads to reduced MAPF efficiency. To address this challenge, this letter introduces a unique reward shaping technique based on Independent Q-Learning (IQL). The aim of this method is to evaluate the influence of one agent on its neighbors and integrate such an interaction into the reward function, leading to active cooperation among agents. This reward shaping method facilitates cooperation among agents while operating in a distributed manner. The proposed approach has been evaluated through experiments across various scenarios with different scales and agent counts. The results are compared with those from other state-of-the-art (SOTA) planners. The evidence suggests that the approach proposed in this letter parallels other planners in numerous aspects, and outperforms them in scenarios featuring a large number of agents.
\end{abstract}


\section{Introduction}
\begin{figure}[t]
\centering
    \includegraphics[width=\linewidth]{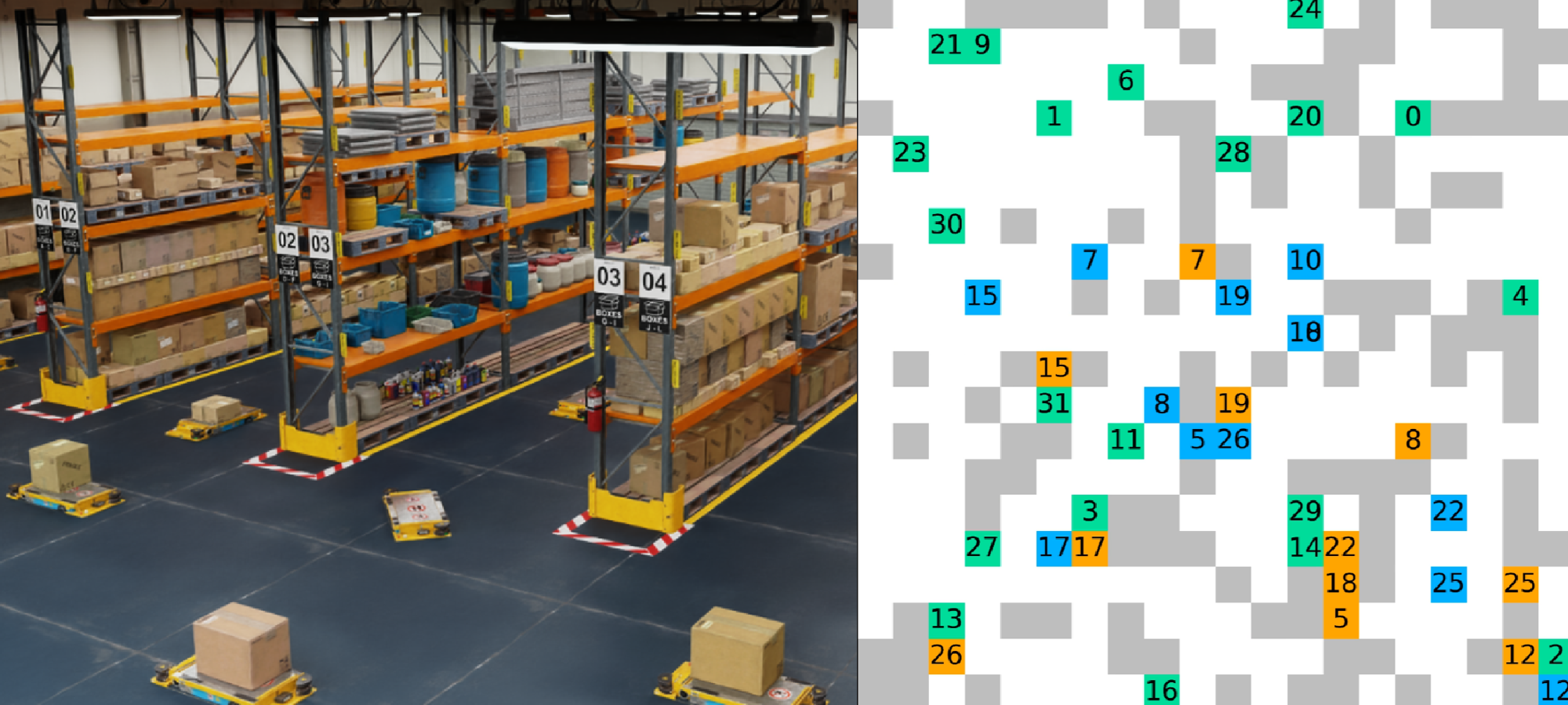}
    \vspace{-0.3cm}
    \caption{Multi-agent scenarios in real-world and grid map with numerous mobile robots. In the grid map, blue cells denote agents, yellow cells indicate their goals, and green cells signify agents that have reached their goals.}
    \label{overview}
    \vspace{-0.3cm}
\end{figure}

\IEEEPARstart{M}{APF} is a fundamental research area in the field of multi-agent systems, aiming to find conflict-free routes for each agent. This has notable implications in various environments such as ports, airports \cite{7867448, wang2021real}, and warehouses \cite{salzman2020research, Yalcin2018, nagorny2012service}. In these scenarios, there are typically a large number of mobile agents. These scenarios can generally be abstracted into grid maps, as illustrated in Fig.~\ref{overview}. The MAPF methodology is divided mainly into two classes: centralized and decentralized algorithms. Centralized algorithms \cite{ferner2013odrm, guo2022sub, zhang2022multi} offer efficient paths by leveraging global information but fall short in scalability when handling a significant number of agents because of increased computational needs and extended planning time. In contrast, decentralized algorithms \cite{wagner2015subdimensional, fan2021decentralized} show better scalability in large-scale environments but struggle to ensure sufficient cooperation among agents, thereby affecting the success rate in pathfinding and overall efficiency.

MARL techniques, particularly those utilizing distributed execution, provide effective solutions to MAPF problems. By modeling MAPF as a partially observable Markov decision process (POMDP), MARL algorithms can develop policies that make decisions based on agents' local observations and inter-agent communication. Given that MARL-trained policy networks do not rely on global observations, these methods exhibit excellent scalability and flexibility in dynamic environments. MARL enhances the success rate and robustness of path planning, making it particularly suitable for large-scale multi-agent scenarios. 
Algorithms such as \cite{peng2012vdn, rashid2020monotonic, foerster2018counterfactual} utilize a centralized training distributed execution (CTDE) framework and foster cooperation between agents using global information during training. However, they struggle to scale for larger numbers of agents due to increasing training costs. In contrast, algorithms based on distributed training distributed execution (DTDE) frameworks \cite{SUTTLE20201549, tampuu2017multiagent} perform well in large-scale systems. However, due to the lack of global information, individual agents tend to solely focus on maximizing their own rewards, resulting in limited cooperation among the whole system.
To address this issue, recent work \cite{chen2023stas, li2021shapley, peng2021learning} emerges that improves the performance of RL algorithms in distributed training frameworks through reward shaping. However, some of these reward shaping methods are too computationally complex, while others lack stability. This instability arises because an agent's rewards are influenced by the actions of other agents, which are typically unknown to this agent.
 
In this letter, a reward shaping method named \textbf{Co}operative \textbf{R}eward \textbf{S}haping (CoRS) is devised to enhance MAPF efficiency within a DTDE framework. The approach is straightforward and tailored for a limited action space. The cooperative trend of action $a^i$ is represented by the maximum rewards that neighboring agents can achieve after agent $A^i$ performs $a^i$. Specifically, when agent $A^i$ takes action $a^i$, its neighbor $A^j$ traverses its action space, determining the maximum reward that $A^j$ can achieve given the condition of $A^i$ taking action $a^i$. The shaped reward is then generated by weighting this index with the reward earned by $A^i$ itself.

The principal contributions of our work are as follows:
\begin{enumerate}
\item This letter introduces a novel reward-shaping method \textbf{CoRS}, designed to promote cooperation among agents in MAPF tasks within the IQL framework. This reward-shaping method is unaffected by actions from other agents, ensures easy convergence during training, and is notable for its computational simplicity.

\item It can be demonstrated that the CoRS method can alter the agent's behavior, making it more inclined to cooperate with other agents, thereby improving the overall efficiency of the system.

\item The CoRS method is challenged against present SOTA algorithms, showcasing equivalent or superior performance, eliminating the necessity for complex network structures.
\end{enumerate}

The rest of this letter is organized as follows: Section~\ref{Related Works} reviews the related works on MAPF. Preliminary concepts are presented in Section~\ref{Preliminary}. The proposed algorithm is detailed in Section~\ref{Approach}. Section~\ref{Experiments} discusses the experimental results. Finally, Section~\ref{Conclusion} concludes this letter.

\section{Related Works} \label{Related Works}

\subsection{MAPF Based on Reinforcement Learning}
RL-based planners such as \cite{wang2020mobile, sartoretti2019primal, liu2020mapper, li2020graph}, typically cast MAPF as a MARL problem to learn distributed policies for agents from partial observations. This method is particularly effective in environments populated by a large number of agents. Techniques like Imitation Learning (IL) often enhance policy learning during this process. Notably, the PRIMAL algorithm \cite{sartoretti2019primal} utilizes the Asynchronous Advantage Actor Critic (A3C) algorithm and applies behavior cloning for supervised RL training using experiences from the centralized planner ODrM* \cite{6631119}. However, the use of a centralized planner limits its efficiency, as solving the MAPF problem can be time-intensive, particularly in complex environments with a large number of agents.

In contrast, Distributed Heuristic Coordination (DHC) \cite{ma2021distributed} and Decision Causal Communication (DCC) \cite{9665278} algorithms do not require a centralized planner. Although guided by an individual agent's shortest path, DHC innovatively incorporates all potential shortest path choices into the model's heuristic input rather than obligating an agent to a specific path. Additionally, DHC collects data from neighboring agents to inform its decisions and employs multi-head attention as a convolution kernel to calculate interactions among agents. DCC is an efficient algorithm that enhances the performance of agents by enabling selective communication with neighbors during both training and execution. Specifically, a neighboring agent is deemed significant only if its presence instigates a change in the decision of the central agent. The central agent only needs to communicate with its significant neighbors. 

\subsection{Reward Shaping}

The reward function significantly impacts the performance of RL algorithms. Researchers persistently focus on designing reward functions to optimize algorithm learning efficiency and agent performance. A previous study \cite{ng1999policy} analyzes the effect of modifying the reward function in Markov Decision Processes on optimal strategies, indicating that the addition of a transition reward function can boost the learning efficiency. In multi-agent systems, the aim is to encourage cooperation among agents through appropriate reward shaping methods, thereby improving overall system performance. \cite{peysakhovich2017prosocial} probes the enhancement of cooperative agent behavior within the context of a two-player Stag Hunt game, achieved through the design of reward functions.  Introducing a prosocial coefficient, the study validates through experimentation that prosocial reward shaping methods elevate performance in multi-agent systems with static network structures. Moreover, \cite{jaques2019social} promotes cooperation among agents through the reward of an agent whose actions causally influence the behavior of other agents. The evaluation of causal impacts is achieved through counterfactual reasoning, with each agent simulating alternative actions at each time step and calculating their effect on other agents' behaviors. Actions that lead to significant changes in the behavior of other agents are deemed influential and are rewarded accordingly.

Several studies, including \cite{chen2024stas, li2021shapley, han2022stable}, utilize the Shapley value decomposition method to calculate or redistribute each agent's cooperative benefits. \cite{han2022stable} confirms that if a transferable utility game is a convex game, the MARL reward redistribution, based on Shapley values, falls within the core, thereby securing stable and effective cooperation. Consequently, agents should maintain their partnerships or collaborative groups. This concept is the basis for a proposed cooperative strategy learning algorithm rooted in Shapley value reward redistribution. The effectiveness of this reward shaping method in promoting cooperation among agents, specifically within a basic autonomous driving scenario, is demonstrated in the paper. However, the process of calculating the Shapley value can be intricate and laborious. Ref.~\cite{chen2024stas} aims to alleviate these computational challenges by introducing approximation of marginal contributions and employing Monte Carlo sampling to estimate Shapley values. Coordinated Policy Optimization (CoPO)\cite{peng2021learning} puts forth the concept of ``cooperation coefficient", which shapes the reward by taking a weighted average of an agent's individual rewards and the average rewards of its neighboring agents, based on the cooperation coefficient. This approach proves that rewards shaped in this manner fulfill the Individual Global Max (IGM) condition. Findings from traffic simulation experiments further suggest that this method of reward shaping can significantly enhance the overall performance and safety of the system. However, this approach ties an agent's rewards not merely to its personal actions but also to those of its neighbors. Such dependencies might compromise the stability during the training process and the efficiency of the converged strategy.

\section{Preliminary} \label{Preliminary}
\subsection{Cooperative Multi-Agent Reinforcement Learning}
Consider a Markov process involving $n$ agents $\{A^1,\dots,A^n\} := \mathbb{A}$, represented by the tuple $(\mathcal{S}, \mathcal{A}, O, \mathcal{R}, \mathcal{P}, \gamma)$. $A^i$ represents agent $i$. At each time step $t$, $A^i$ chooses an action $a^i_t$ from its action space $\mathcal{A}^i$ based on its state $s^i_t \in \mathcal{S}^i$ and observation $o^i \in O$ according to its policy $\pi^i$. All $a^i_t$ form a joint action $\bar{a}_t = \{a^1_t, \dots, a^n_t\} \in (\mathcal{A}^1 \times \dots \times \mathcal{A}^n) := \mathcal{A}$, and all $s^i_t$ form a joint state $\bar{s}_t = \{s^1_t, \dots, s^n_t\} \in (\mathcal{S}^1 \times \dots \times \mathcal{S}^n) := \mathcal{S}$. For convenience in further discussions, agent's local observations $o^i_t$ are treated as a part of the agent’s state $s^i_t$. Whenever a joint action $\bar{a}_t$ is taken, the agents acquire a reward $\bar{r} = \{r^1_t, r^2_t, ..., r^n_t\} \in \mathcal{R}$, which is determined by the local reward function $r^i_t(\bar{s}_t, \bar{a}_t): \mathcal{S} \times \mathcal{A} \rightarrow \mathbb{R}$ with respect to the joint state and action. The state transition function $\mathcal{P}(\bar{s}_{t+1} | \bar{s}_t, \bar{a}_t): \mathcal{S} \times \mathcal{S} \times \mathcal{A} \rightarrow [0, 1]$ characterizes the probability of transition from the current state $\bar{s}_t$ to $\bar{s}_{t+1}$ under action $\bar{a}_t$. The policy $\pi^i(a^i_t | s^i_t)$ provides the probability of $A^i$ taking action $a^i_t$ in the state $s^i_t$. $\bar{\pi}$ represents the joint policy for all agents. The action-value function is given by $Q^i_\pi(s^i_t, a^i_t) = \mathbb{E}_{\tau^i \sim \pi^i} [\sum_{t=0}^{T} \gamma^t r^i_t]$, where the trajectory $\tau^i = [s^i_{t+1}, a^i_{t+1}, s^i_{t+2}, ...]$ represents the path taken by the agent $A^i$. 
The state-value function is given by $V^i_{\pi}(s) = \sum_{a \in \mathcal{A}^i} \pi(a | s^i_t) Q^i_{\pi}(s^i_t, a)$ and the discounted cumulative reward is $J^i = \mathbb{E}_{s_0 \sim \rho_0} V^i_{\pi}(s_0)$,  where $\rho_0$ represents the initial state distribution. For cooperative MARL tasks, the objective is to maximize the total cumulative reward $J^{tot} = \sum_{i = 1}^n J^i$ for all agents.

\subsection{Multi-agent Pathfinding Environment Setup}
This letter adopts the same definition of the multi-agent path-finding problem as presented in \cite{ma2021distributed, 9665278}.
Consider $n$ agents 
in an $w \times h$ undirected grid graph $G(V,E)$ with $m$ obstacles $\{B^1,\dots,B^m\}$, where $V = \{v(i,j)|1 \leq i \leq w, 1 \leq j \leq h\}$ is the set of vertices in the graph, and all agents and obstacles located within $V$. All vertices $v(i,j) \in V$ follow the 4-neighborhood rule, that is, $[v(i,j), v(p, q)] \in E$ for all $v(p,q) \in \{v(i, j \pm 1), v(i \pm 1, j)\} \cap V$. Each agent $A^i$ has its unique starting vertex $s^i$ and goal vertex $g^i$, and its position at time $t$ is $x^i_t \in V$. The position of the obstacle $B^k$ is represented as $b^k \in V$. At each time step, each agent can execute an action chosen from its action space $\mathcal{A}^i = \{\text{``Up'', ``Down'', ``Left'', ``Right'', ``Stop''}\}$. During the execution process, two types of conflict can arise: vertex conflict ($x^i_t = x^j_t$ or $x^i_t = b^k$) and edge conflict ($[x^i_{t-1}, x^i_{t}] = [x^j_t, x^j_{t-1}]$).  If two agents conflict with each other, their positions remain unchanged. 
The subscript $t$ for all the aforementioned variables can be omitted as long as it does not cause ambiguity. The goal of the MAPF problem is to find a set of non-conflicting paths $\bar{P} = \{P^1, P^2, ..., P^n\}$ for all agents, where the agent's path $P^i=[s^i, x^i_1, \dots, x^i_t, \dots, g^i]$ is an ordered list of $A^i$'s position. Incorporating the setup of multi-agent reinforcement learning, we design a reward function for the MAPF task, as detailed in Table~\ref{reward function}. The design of the reward function basically follows \cite{ma2021distributed, 9665278}, with slight adjustments to increase the reward for the agent moving towards the target to better align with our reward shaping method.
\begin{table}[t]
\begin{center}
    \caption{Reward Function.}
    \label{reward function}
    \vspace{-0.1cm}
    \begin{tabular}{ c | c }
    \toprule
    \hline
    Action & Reward\\
    \hline
    Move (towards goal, away from goal) & -0.070, -0.075 \\
    \hline
    Stay (on goal, off goal) & 0, -0.075 \\ 
    \hline
    Collision (obstacle/agents) & -0.5\\
    \hline 
    Finish & 3 \\
    \hline
    \bottomrule
    \end{tabular}
    \end{center}
\end{table}

\section{Cooperative Reward Shaping} \label{Approach}
\begin{figure*}[t] 
    \includegraphics[width=\linewidth]{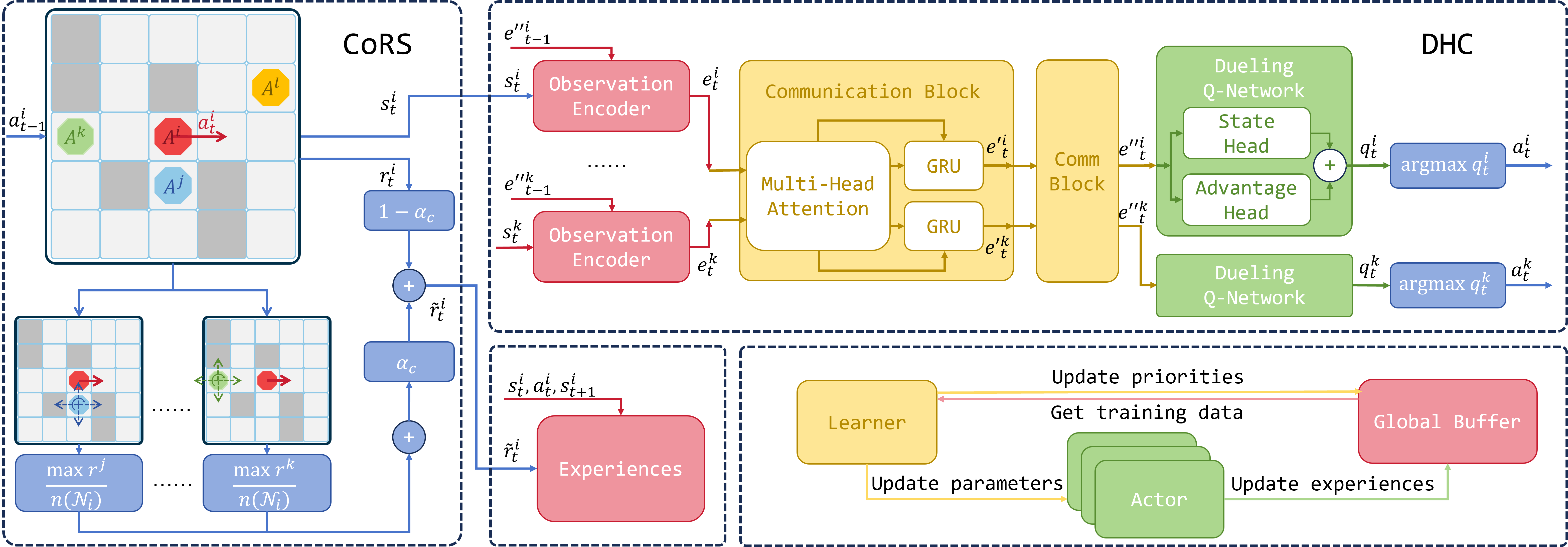}
    \vspace{-0.7cm}
    \caption{The combined framework of CoRS and DHC algorithms. The CoRS component predominantly shapes rewards. DHC component comprises of communication blocks and dueling Q networks. Notably, the communication block employs a multi-head attention mechanism. The framework utilizes parallel training as an efficient solution. This process involves the simultaneous generation of multiple actors to produce experiential data and upload it to the global buffer. The learner then retrieves this data from the global buffer for training purposes, thereby allowing for frequent updates to the Actor's network.}
    \label{framework-pic}
\end{figure*}

Many algorithms employing MARL techniques to address the MAPF problem utilize IQL or other decentralized training and execution frameworks to ensure good scalability. For example, \cite{ma2021distributed, 9665278} are developed based on IQL.
Although IQL can be applied to scenarios with a large number of agents, it often performs poorly in tasks that require a high degree of cooperation among agents, such as the MAPF task. This poor performance arises because, within the IQL framework, each agent greedily maximizes its own cumulative reward, leading agents to behave in an egocentric and aggressive manner, thus reducing the overall efficiency of the system.
To counteract this, this letter introduces a reward shaping method named \textbf{Co}operative \textbf{R}eward \textbf{S}haping (CoRS), and combines CoRS with the DHC algorithm. The framework combining CoRS with DHC is shown in Fig.~\ref{framework-pic}. The aim of CoRS is to enhance performance within MAPF problem scenarios. The employment of reward shaping intends to stimulate collaboration among agents, effectively curtailing the performance decline in the multi-agent system caused by selfish behaviors within a distributed framework.

\subsection{Design of the Reward Shaping Method}
In the MAPF task, the policies trained using IQL often result in scenarios where one agent blocks the path of other agents or collides with them. To enhance cooperation among agents within the IQL framework, a feasible approach is reward shaping. Reward shaping involves meticulously designing the agents' reward functions to influence their behavior. For example, when a certain type of behavior needs to be encouraged, a higher reward function is typically assigned to that behavior. Thus, to foster cooperation among agents in the MAPF problem, it is necessary to design a metric that accurately evaluates the collaboration of agents' behavior and incorporate this metric into the agents' rewards. Consequently, as each agent maximizes its own reward, it will consider the impact of its action on other agents, thereby promoting cooperation among agents and improving overall system efficiency.

Let $I^i_c(\bar{s}_t, \bar{a}_t)$ be a metric that measures the cooperativeness of the action $a^i_t$ of agent $A^i$. To better regulate the behavior of the agent, \cite{peysakhovich2017prosocial} introduces a cooperation coefficient $\alpha$ and shapes the agent's reward function in the following form:
\begin{equation}
    \label{general rs method}
    \tilde{r}^i_t = (1 - \alpha) r^i_t + \alpha I^i_c(\bar{s}_t, \bar{a}_t),
\end{equation}
where $\alpha$ describes the cooperativeness of the agent. When $\alpha = 0$, the agent completely disregards the impact of its actions on other agents, acting entirely selfishly and when $\alpha = 1.0$, the agent behaves with complete altruism.
For agent $A^i$, an intuitive approach to measure the cooperativeness of $A^i$'s behavior is to use the average reward of all agents except $A^i$:
\begin{equation}
\label{original Ic}
    I^i_c(\bar{s}_t, \bar{a}_t) = \frac{1}{|A^{-i}|} \sum_{j \in A^{-i}} r^j(\bar{s}_t, \bar{a}_t),
\end{equation}
where $A^{-i}$ denotes the set of all agents except $A^i$, and $|A^{-i}|$ represents the number of agents in $A^{-i}$.
This reward shaping method is equivalent to the \textit{neighborhood reward } proposed in \cite{peng2021learning} when $d_n$, the neighborhood radius of the agent, approaches infinity.
The physical significance of Eq.~\eqref{original Ic} is as follows: If the average reward of the agents other than $A^i$ is relatively high, it indicates that $A^i$'s actions have not harmed the interests of other agents. Hence, $A^i$'s behavior can be considered as cooperative. Conversely, if the average reward of the other agents is low, it suggests that $A^i$'s behavior exhibits poor cooperation.

\begin{figure}[t]
    \includegraphics[width=\linewidth]{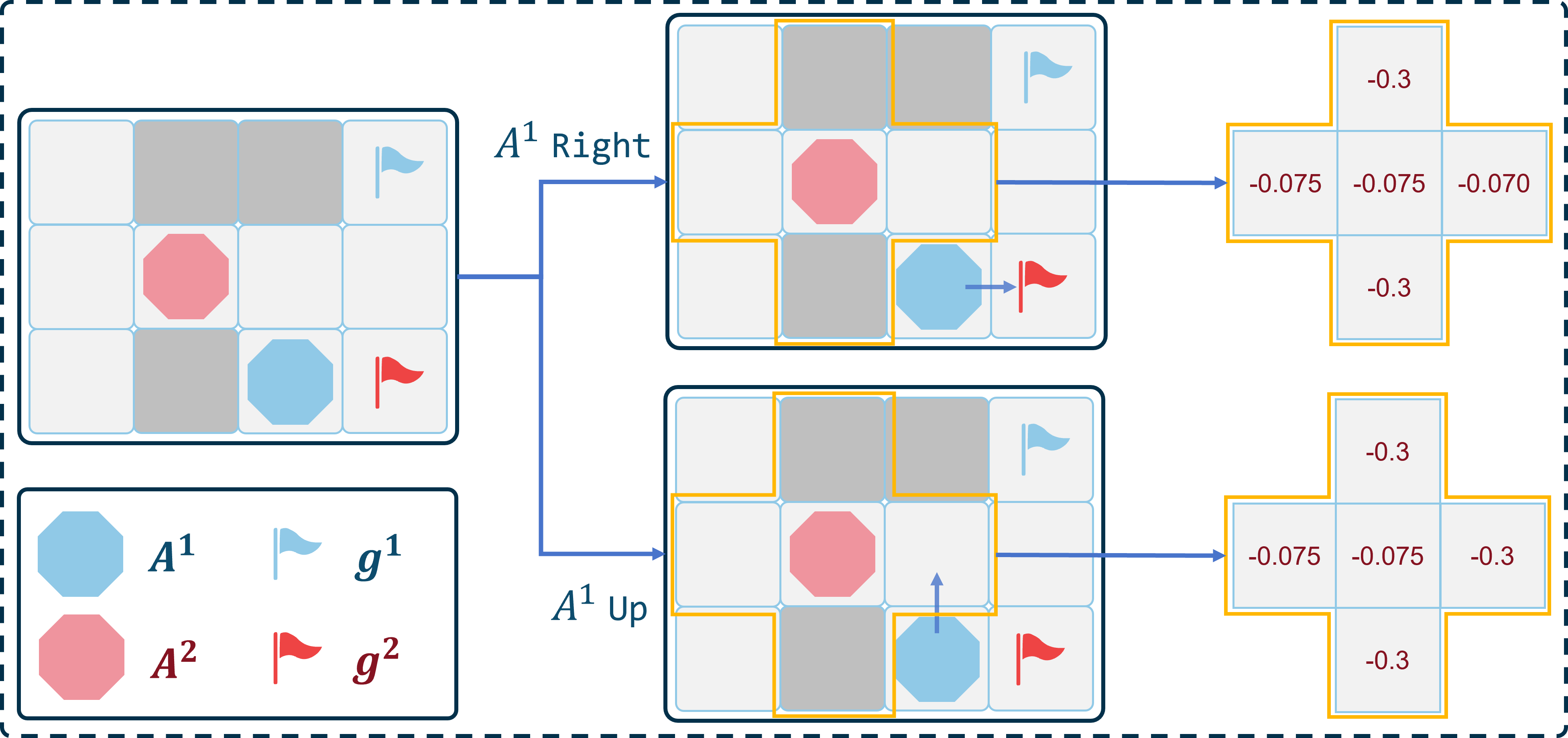}
    \vspace{-0.7cm}
    \caption{An example of the reward shaping method.}
    \label{example 1}
    \vspace{-0.2cm}
\end{figure}

However, $I^i_c(\bar{s}_t, \bar{a}_t)$ in Eq.~\eqref{original Ic} is unstable, which is not only related to $a^i_t$ but is also strongly correlated with the actions $a^{-i} = \{a^j | A^j \in \mathbb{A}, j \ne i\}$ of other agents. Appendix~\ref{subsection:instability} provides specific examples to illustrate this instability. Within the IQL framework, this instability in the reward function makes learning of Q-values challenging and can even prevent convergence. 
To address this issue, this letter proposes a new metric to assess the cooperativeness of agent behavior. The specific form of this metric is as follows:
\begin{equation}
    \label{my Ic}
    I^i_c(\bar{s}_t, \bar{a}_t) = \max_{a^{-i}} \sum_{j \in  A^{-i}} \frac{r^j(\bar{s}_t, \{a^i_t, a^{-i}\})}{|A^{-i}|}.
\end{equation}
Here $\bar{a}_t = \{a^i_t, a^{-i}_t\}$. The use of the $\max$ operator in Eq.~\eqref{my Ic} eliminates the influence of $a^{-i}_t$ on $I^i_c$ while reflecting the impact of $a^i_t$ on other agents. The term $\max_{a^{-i}} \sum_{j \in A^{-i}} r^j(\bar{s}_t, \{a^i_t, a^{-i}\})$ represents the maximum reward that all the agents except $A^i$ can achieve under the condition of $\bar{s}_t$ and $a^i_t$, whereas the actual value of $\sum_{j \in A^{-i}} r^j(\bar{s}_t, \{a^i_t, a^{-i}_t\})$ is determined by $a^{-i}_t$ when $a^i_t$ is given. Accordingly, it holds true under any circumstances that $\sum_{j \in A^{-i}} r^j(\bar{s}_t, \{a^i_t, a^{-i}_t\}) \leq \max_{a^{-i}} \sum_{j \in A^{-i}} r^j(\bar{s}_t, \{a^i_t, a^{-i}\})$. 
The complete reward shaping method is then as follows:
\begin{equation}
    \label{reward shaping method}
    \tilde{r}^i_t(\bar{s}_t, a^i_t) = (1 - \alpha) r^i_t + \alpha \max_{a^{-i}} \sum_{j \in A^{-i}} \frac{r^j(\bar{s}_t, \{a^i_t, a^{-i}\})}{|A^{-i}|}.
\end{equation}
Fig.~\ref{example 1} gives an one step example to illustrate how this reward-shaping approach Eq.~\eqref{reward shaping method} is calculated and promotes inter-agent cooperation when $\alpha = \frac{1}{2}$. 
There are two agents, $A^1$ and $A^2$, along with some obstacles. Without considering $A^2$, $A^1$ has two optimal actions: ``Up'' and ``Right''. Fig.~\ref{example 1} also illustrates the potential rewards that $A^2$ may receive when $A^1$ takes different actions. 
Table~\ref{reward shaping details} presents the specific reward values for each agent in this example.
According to the reward shaping method proposed in Eq.~\eqref{reward shaping method}, choosing the ``Right'' action could yield a higher one-step reward, and in such case, $A^2$ should also take action``Right''.
Appendix \ref{appendix:B-C} provides another analysis of this example from the vantage point of the cumulative rewards of the agent, further elucidating how the reward shaping method Eq.~\eqref{reward shaping method} facilitates cooperation among agents.


\begin{table}[t]
    \caption{Rewards for Agents Under Different Actions.} \label{reward shaping details}
    \setlength{\abovecaptionskip}{-0.4cm} \label{tab2}
    \vspace{-0.1cm}
    \renewcommand{\arraystretch}{1.3}
    \centering
    \begin{tabular}{c|c|c|c|c|c|c|c} 
        \toprule
        \hline
        \multirow{2}{*}{$a^1$} & \multicolumn{5}{c|}{$a^2$} & \multirow{2}{*}{$\max r^2$} & \multirow{2}{*}{$\tilde{r}^1$} \\ \cline{2-6}
                      & $\uparrow$ & $\downarrow$ & $\leftarrow$ & $\rightarrow$ & wait   & &                \\ \hline
        $\rightarrow$ & -0.3       & -0.3         & -0.075       & -0.070        & -0.075 & -0.07  & -0.140  \\
        $\uparrow$    & -0.3       & -0.3         & -0.075       & -0.3          & -0.075 & -0.075 & -0.145  \\ \hline
        \bottomrule
    \end{tabular}
\end{table}



\subsection{Analysis of Reward Shaping}
This section provides a detailed analysis of how reward shaping method Eq.~\eqref{reward shaping method} influences agent behavior. For ease of discussion, the subsequent analysis will be conducted from the perspective of $A^i$. 


In the MAPF problem, the actions of agents are tightly coupled. The action of $A^i$ may impact $A^j$, and subsequently, the action of $A^j$ may also affect $A^k$. Thus, the action of $A^{i}$ indirectly affect $A^{k}$.
This coupling makes it challenging to analyze the interactions among multiple agents. To mitigate this coupling, we consider all agents in $A^{-i}$ as a single virtual agent $\tilde{A}^{-i}$. The action of $\tilde{A}^{-i}$ is $a^{-i}$ and its reward $r^{-i}$ is their average reward $\frac{1}{|A^{-i}|} \sum_{j \in A^{-i}} r^j$. The use of average here ensures that $A^i$ and $\tilde{A}^{-i}$ are placed on an equal footing. The virtual agent $\tilde{A}^{-i}$ must satisfy the condition that no collisions occur between the agents constituting $\tilde{A}^{-i}$.
It is important to note that the interaction between $A^i$ and $\tilde{A}^{-i}$ is not entirely equivalent to the interaction between $A^i$ and agents in $A^{-i}$. 
This is because when considering the agents in $A^{-i}$ as $\tilde{A}^{-i}$, all robots within $\tilde{A}^{-i}$ fully cooperate. In contrast, treating these agents as independent individuals makes it difficult to ensure full cooperation.
Nonetheless, $\tilde{A}^{-i}$ can still represent the ideal behavior of agents in $A^{-i}$. Therefore, analyzing the interaction between $A^i$ and $\tilde{A}^{-i}$ can still illustrate the impact of reward shaping on agent interactions.
Consider the interaction between $A^i$ and $\tilde{A}^{-i}$ in the time period $[t_s, t_e]$. 
$\mathcal{Q}_{\pi}^i (\bar{s}, a^i) = \mathbb{E}_{\tau} \left[ \sum\limits_{t=t_s}^{t_e} \gamma^{t - t_s} \tilde{r}^i(\bar{s}_t, a^i_t) \bigg| \bar{s}_{t_s} = \bar{s}, a^i_{t_s} = a^i \right]$ and $\mathcal{Q}_{\pi}^{-i}(\bar{s}, a^{-i}) = \mathbb{E}_{\tau} \left[ \sum\limits_{t=t_s}^{t_e} \gamma^{t - t_s} \tilde{r}^{-i}(\bar{s}_t, a^{-i}_t) \bigg| \bar{s}_{t_s} = \bar{s}, a^{-i}_{t_s} = a^{-i} \right]$ are the cumulative reward of $A^i$ and $\tilde{A}^{-i}$. 
$Q^{tot} (\bar{s}, \bar{a}) = \mathbb{E}_{\tau} \left[ \sum\limits_{t=t_s}^{t_e} \gamma^{t - t_s} (r^i + r^{-i}) \bigg| \bar{s}_{t_s} = \bar{s}, \bar{a}_{t_s} = \bar{a} \right]$ is the cumulative reward for both $A^i$ and $\tilde{A}^{-i}$.
The optimal policies $\pi^i_* = \operatorname{argmax}_{\pi} \mathcal{Q}_{\pi}^i (\bar{s}, a^i)$, $\pi^{-i}_* = \operatorname{argmax}_{\pi} \mathcal{Q}_{\pi}^{-i} (\bar{s}, a^{-i})$, and $\bar{\pi}_* = \operatorname{argmax}_{\pi} Q^{tot}_{\pi} (\bar{s}, \bar{a})$. 
$A^i$ and $\tilde{A}^{-i}$ will select their actions according to $a^i_t = \operatorname{argmax}_{a^i} \mathcal{Q}_{\pi}^i (\bar{s}_t, a^i)$ and $a^{-i}_t = \operatorname{argmax}_{a^{-i}} \mathcal{Q}_{\pi}^{-i} (\bar{s}_t, a^{-i})$. We hope that $a^i_t$ and $a^{-i}_t$ will collectively maximize $Q^{tot} (\bar{s}, \bar{a}) = Q^{tot} (\bar{s}, \{a^{i}_t, a^{-i}_t\})$. Specifically,
\begin{equation*}
    \operatorname{argmax}_{\bar{a}}  Q^{tot}_{\bar{\pi}_*}(\bar{s}, \bar{a}) = 
    \begin{Bmatrix}
    \operatorname{argmax}_{a^i} \mathcal{Q}_{\pi_*^i}^i (\bar{s}, a^i),\\
    \operatorname{argmax}_{a^{-i}} \mathcal{Q}_{\pi_*^{-i}}^{-i} (\bar{s}, a^{-i})
    \end{Bmatrix}.
\end{equation*}
That is, $\mathcal{Q}_{\pi_*^i}^i$, $\mathcal{Q}_{\pi_*^{-i}}^{-i}$, and $Q^{tot}_{\bar{\pi}_*} (\bar{s}, \bar{a})$ satisfy the \textit{Individual-Global-Max (IGM) condition}. The definition of the \textit{IGM condition} can be found in \cite{son2019qtran}.
We introduce the following two assumptions to facilitate the analysis.
\begin{assumption}
    \label{assumption 1}
    The reward for the agents staying at the target point is $0$, while the rewards for both movement and staying at non-target points are $r_m < 0$, and the collision reward $r_c < r_m$.
\end{assumption}
\begin{assumption}
    \label{assumption 2}
    During the interaction process between $A^i$ and $\tilde{A}^{-i}$, neither $A^i$ nor $\tilde{A}^{-i}$ has ever reached its respective endpoint.
\end{assumption}
Here, $\tilde{A}^{-i}$ not reaching destination means $\forall j \in A^{-i}$, $A^j$ has not reached its destination. 
Based on the assumptions, it can be proven that:
\begin{theorem} \label{Q IGM}
    Assume Assumps.~\ref{assumption 1} and~\ref{assumption 2} hold. Then when $\alpha = \frac{1}{2}$, $\mathcal{Q}^i_{\pi^i_*} (\bar{s}_t, a^i)$, $\mathcal{Q}^{-i}_{\pi^{-i}_*} (\bar{s}_t, a^{-i})$ and $Q^{tot}_{\bar{\pi}_*} (\bar{s}_t, \{a^i, a^{-i}\})$ satisfy the \textit{IGM condition}.
\end{theorem}

\begin{proof}
    The proof is provided in Appendix~\ref{proofs}.
\end{proof}

Theorem~\ref{Q IGM} demonstrates that the optimal policy, trained using the reward function given by Eq.~\eqref{reward shaping method}, maximizes the overall rewards of $A^i$ and the virtual agents $\tilde{A}^{-i}$, rather than selfishly maximizing its own cumulative reward. This implies that $A^i$ will actively cooperate with other agents, thus improving the overall efficiency of the system.
It should be noted that the impact of $\alpha$ on the agent's behavior is complex.
The choice of $\alpha = \frac{1}{2}$ here is a specific result derived from using the virtual agent $\tilde{A}^{-i}$. 
Although this illustrates how the reward shaping method in Eq.~\eqref{reward shaping method} induces cooperative behavior in agents, it does not imply that $\frac{1}{2}$ is the optimal value of $\alpha$ in all scenarios.

\subsection{Approximation}
Theorem \ref{Q IGM} illustrates that the reward shaping method Eq.~\eqref{reward shaping method} can promote cooperation among agents in MAPF tasks. However, calculating Eq.~\eqref{my Ic} requires traversing the joint action space of $A^{-i}$. For the MAPF problem  where the action space size for each agent is 5, the joint action space for $n$ agents contains up to $5^n$ states, significantly reducing the computational efficiency of the reward shaping method.
\begin{wrapfigure}{r}{.33\linewidth} 
    \vspace{-0.4cm}
    \includegraphics[width=\linewidth]{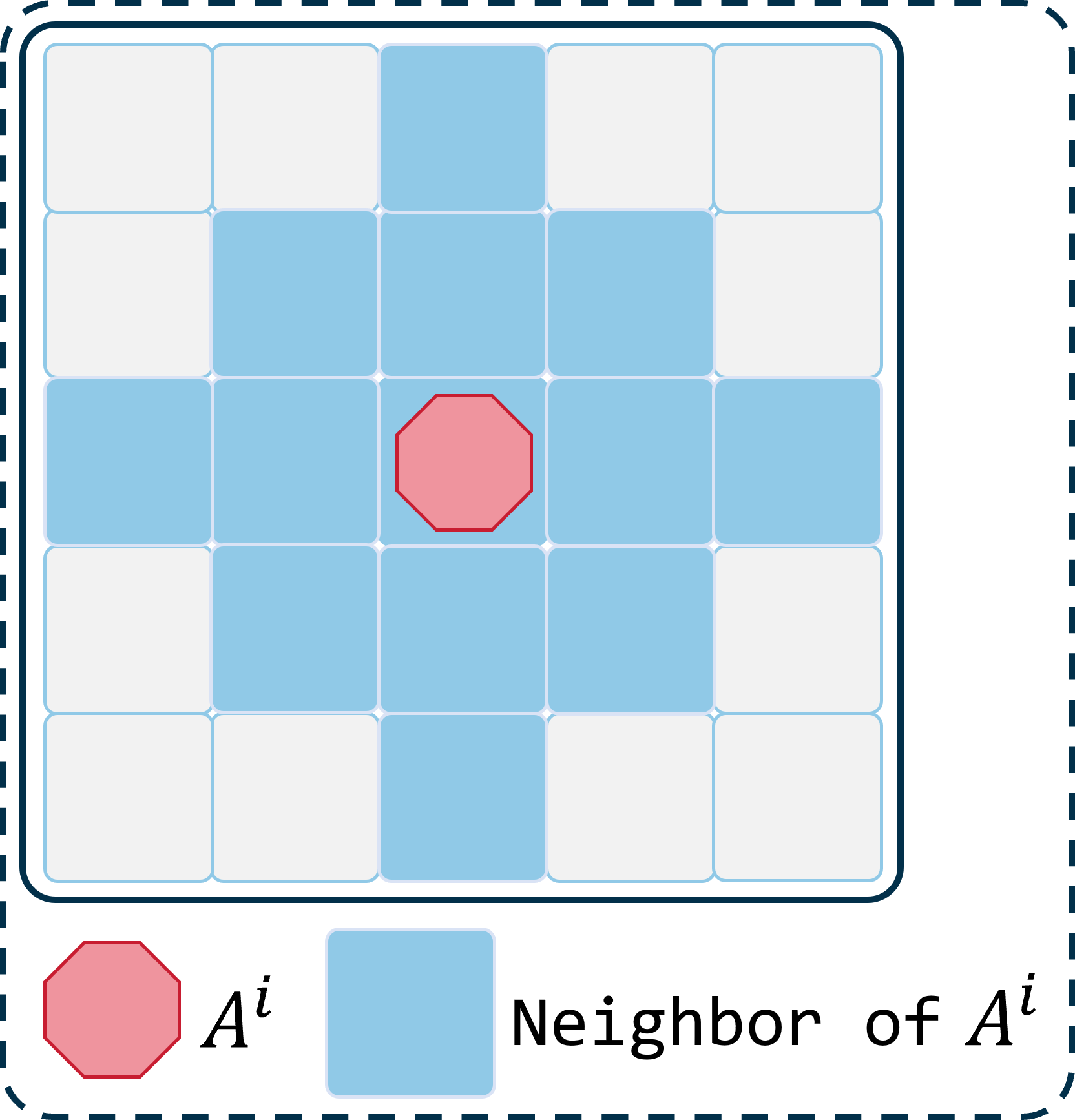}
    \vspace{-0.7cm}
    \caption{Neighbors of $A^i$ when $d_n = 2$.} 
    \vspace{-0.5cm}
    \label{fig:tibet}
\end{wrapfigure}
Therefore, we must simplify the calculation of Eq.~\eqref{my Ic}. Before approximating Eq.~\eqref{my Ic}, we first define the neighbors of an agent in the grid map as follows:
$A^j$ is considered a neighbor of $A^i$ if the Manhattan distance between them is no more than $d_n$, where $d_n$ is the neighborhood radius of the agent.
If the Manhattan distance between two agents is no more than 2, a conflict may arise between them in a single step. Therefore, we choose $d_n = 2$, which maximally simplifies interactions between agents while adequately considering potential collisions. Let $\mathcal{N}_i$ denote the set of neighboring agents of $A^i$, and $|\mathcal{N}_i|$ represent the number of neighbors of $A^i$. Fig.~\ref{fig:tibet} illustrates the neighbors of $A^i$ when $d_n = 2$.

Considering that in MAPF tasks, the direct interactions among agents are constrained by the distances between them, the interactions between any given agent and all other agents can be simplified to the interactions between the agent and its neighbors. That is:
\begin{equation*}
    I^i_c(\bar{s}_t, \bar{a}_t) \approx \max_{a^{\mathcal{N}_i}} \frac{1}{|\mathcal{N}_i|} \sum_{j \in \mathcal{N}_i} r^j(\bar{s}_t, \{a^i_t, a^{\mathcal{N}_i}\}),
\end{equation*}
where $a^{\mathcal{N}_i}$ represents the joint actions of all neighbors of $A^i$. Next, we approximate $\max_{a^{\mathcal{N}_i}} \frac{1}{|\mathcal{N}_i|} \sum_{j \in \mathcal{N}_i} r^j(\bar{s}_t, \{a^i_t, a^{\mathcal{N}_i}\})$ using $\frac{1}{|\mathcal{N}_i|} \sum_{j \in \mathcal{N}_i} \max_{a^j}  r^j(\bar{s}_t, \{a^i_t, a^j\})$. This approximation implies that we consider the interaction between agent $A^i$ and one of its neighbors $A^j$ at a time. 
The computational complexity of the approximate $I_c$ is much lower than the original complexity.

To approximate $\bar{s}_t$, we posit that when $A^i$ can observe its two-hop neighbors, its observation is considered sufficient, and its state $s^i_t$ can approximate $\bar{s}_t$ to a certain extent, i.e., $\tilde{r}^i(\bar{s}_t, a^i_t) = \tilde{r}^i(s^i_t, a^i_t)$. 
When $d_n = 2$, the minimum observation range includes all grids within a Manhattan distance of no more than $2 \cdot d_n = 4$ from the agent.
For convenience, a $9 \times 9$ square is chosen as the agent's field of view (FOV), consistent with the settings in \cite{ma2021distributed} and covering the agent's minimum observation range.
Consequently, the final reward shaping method is given by:
\begin{equation} \label{Reward Shaping}
    \tilde{r}^i(s^i, a^i) = (1 - \alpha) r^i_t + \alpha \frac{1}{|\mathcal{N}_i|} \sum_{j \in \mathcal{N}_i} \max_{a^j \in \mathcal{A}} r^j(s^i_t, \{a^i_t, a^j\}).
\end{equation}

\begin{figure}[t]
    \vspace{-0.4cm}
    \centering
    \includegraphics[width=\linewidth]{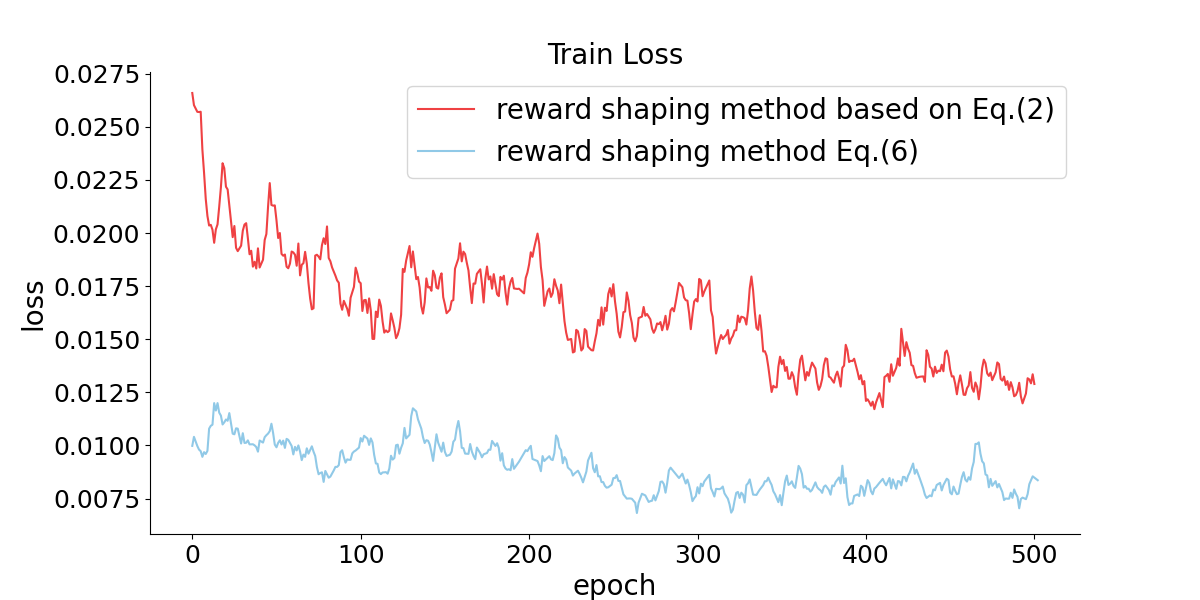}
    \vspace{-0.7cm}
    \caption{Training losses of two different reward shaping methods in 10 $\times$ 10 map with 10 agents. Eq.~\eqref{Reward Shaping} significantly reduces the training loss compared to the method uses Eq.~\eqref{original Ic}.}
    \label{fig:loss}
    \vspace{-0.6cm}
\end{figure}

This reward shaping method possesses the following characteristics:
\begin{enumerate}
    \item $\tilde{r}^i$ for $A^i$ is explicitly dependent on $a^i$ and is explicitly independent of $a^j$ for $j \neq i$. This property improves the stability of the agent's reward function and facilitates the convergence of the network, as shown in Fig.~\ref{fig:loss}.

    \item This method of reward shaping is applicable to MAPF scenarios where the relationship among agents' neighbors is time-varying.

    \item This method, which considers the interaction between just two agents at a time, dramatically reduces the computational difficulty of the reward.
\end{enumerate}

\subsection{Adjustment of the cooperation coefficient} \label{adjustment of cp}
For Eq.~\eqref{Reward Shaping}, it is necessary to adjust $\alpha$ so that the agent can balance its own interests with the interests of other agents.
To find an appropriate cooperation coefficient $\alpha$, it is advisable to examine how the policies trained under different $\alpha$ perform in practice and then employ a gradient descent algorithm based on the performance of the policy to optimize $\alpha$. The cumulative reward $J$ of the agent often serves as a performance metric in RL. However, in environments with discrete action and state spaces, the cumulative reward $J$ is non-differentiable with respect to the $\alpha$, presenting a significant challenge for adjusting $\alpha$. Some work employs zero-order optimization of stochastic gradient estimation to handle non-differentiable optimization problems, that is, estimating the gradient by finite differences of function values. We also use differences in the cumulative reward of the agent to estimate the gradient of the cooperation coefficient $\alpha$:
\begin{equation*}
    \begin{aligned}
        \frac{\hat{\partial} J(\alpha)}{\partial \alpha}  = \frac{J(\alpha + u) - J(\alpha)}{u}, u \in [-\epsilon, \epsilon],
    \end{aligned}
\end{equation*}
where $\epsilon$ represents the maximum step size of the differential. If the update step length is too small, the updates are halted. To expedite the training process, a method of fine-tuning the network is employed. That is, after each adjustment of $\alpha$, the network is not trained from the initial state. Instead, fine-tuning training is performed based on the optimal policy network obtained previously. This training method improves the speed of the training process.

\section{Experiments} \label{Experiments}
We conducted our experiments in the standard MAPF environment, where each agent has a $9 \times 9$ FOV and can communicate with up to two nearest neighbors. Following the curriculum learning method \cite{bengio2009curriculum} used by DHC, we gradually introduced more challenging tasks to the agents. Training began with a simple task that involved a single agent in a $10 \times 10$ environment. Upon achieving a success rate above 0.9, we either added an agent or increased the environment size by 5 to establish two more complex tasks. The model was ultimately trained to handle 10 agents in a $40 \times 40$ environment. The maximum number of steps per episode was set to 256. Training was carried out with a batch size of 192, a sequence length of 20, and a dynamic learning rate starting at $10^{-4}$, which was halved at 100,000 and 300,000 steps, with a maximum of 500,000 training steps. During fine-tuning, the learning rate was maintained at $10^{-5}$. Distributed training was used to improve efficiency, with 16 independent environments running in parallel to generate agent experiences, which were uploaded to a global buffer. The learner then retrieved these data from the buffer and trained the agent's strategy on a GPU. CoRS-DHC adopted the same network structure as DHC. All training and testing were performed on an Intel$^\circledR$ i5-13600KF and Nvidia$^\circledR$ RTX2060 6G.

\subsection{Impact of Reward Shaping}
Following several rounds of network fine-tuning and updates to the cooperation coefficient $\alpha$, a value of $\alpha = 0.1675$ and strategy $\pi$ are obtained. Upon obtaining the final $\alpha$ and $\pi$, the CoRS-DHC-trained policy is compared with the original DHC algorithm-trained policy to assess the effect of the reward shaping method on performance improvement. For a fair comparison, the DHC and CoRS-DHC algorithms are tested on maps of different scales ($40 \times 40$ and $80 \times 80$) with varying agent counts $\{4, 8, 16, 32, 64\}$.  Recognizing the larger environmental spatial capacity of the $80 \times 80$ map, a scenario with 128 agents is also introduced for additional insights. Each experimental scenario includes 200 individual test cases, maintaining a consistent obstacle density of 0.3. The maximum time steps for the $40 \times 40$ and $80 \times 80$ maps are 256 and 386, respectively.

\begin{figure}[t]
    \includegraphics[width=\linewidth]{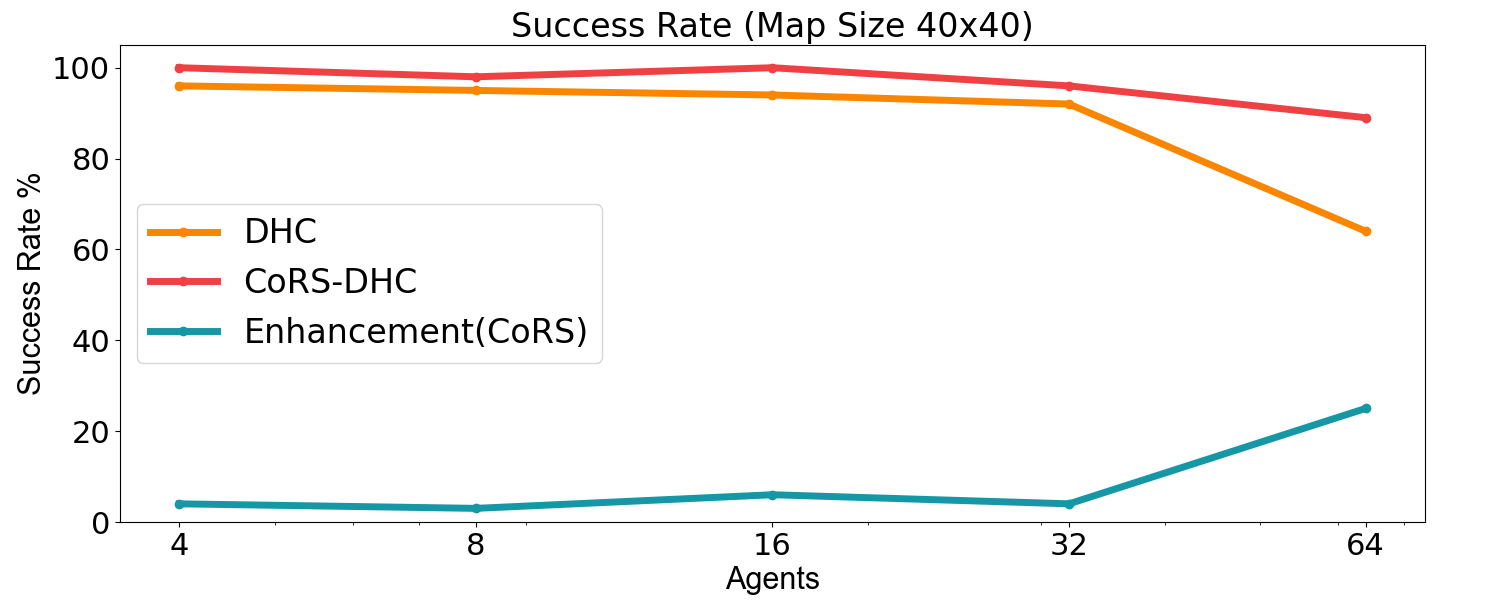}
    \vspace{-0.4cm}
    \caption{The Enhancement of DHC through CoRS. This figure demonstrates the success rate of CoRS-DHC and DHC under different testing scenarios. The table provides the average steps required to complete tasks in different testing environments.}
    \label{improvement}
\end{figure}

\begin{table}[t]
    \vspace{-0.2cm}
    \caption{Average Steps with and without CoRS} \label{improvement-table}
    \setlength{\abovecaptionskip}{-0.4cm} \label{tab2}
    \vspace{-0.1cm}
    \centering
    \begin{tabular}{r|c|c|c|c} 
        \toprule
        \hline
        Steps & \multicolumn{2}{c|}{Map Size 40 $\times$ 40} & \multicolumn{2}{c}{Map Size 80 $\times$ 80}  \\
        \hline
        Agents & DHC    & CoRS-DHC     & DHC    & CoRS-DHC     \\
        \hline
        4      & 64.15  & \red{50.36}  & 114.69 & \red{92.14}  \\
        8      & 77.67  & \red{64.77}  & 133.39 & \red{109.15} \\
        16     & 86.87  & \red{68.48}  & 147.55 & \red{121.25} \\
        32     & 115.72 & \red{95.42}  & 158.58 & \red{137.06} \\
        64     & 179.69 & \red{151.02} & 183.44 & \red{153.06} \\
        128    &        &              & 213.75 & \red{193.50} \\
        \bottomrule
    \end{tabular}
\end{table}

    


From the experimental results illustrated in Fig.~\ref{improvement}, it is evident that our proposed CoRS-DHC significantly outperforms the existing DHC algorithm in all test sets. In high agent density scenarios, such as $40 \times 40$ with 64 agents and $80\times 80$ with 128 agents, our CoRS-DHC improves the pathfinding success rate by more than 20\% compared to DHC. Furthermore, as shown in Table~\ref{improvement-table}, CoRS effectively reduces the number of steps required for all agents to reach their target spot in various test scenarios. Although the DHC algorithm incorporates heuristic functions and inter-agent communication, it still struggles with cooperative issues among agents. In contrast, our CoRS-DHC promotes inter-agent collaboration through reward shaping, thereby substantially enhancing the performance of the DHC algorithm in high-density scenarios. Notably, this significant improvement was achieved without any changes to the algorithm's structure or network scale, underscoring the effectiveness of our approach.

\subsection{Success Rate and Average Step}
Additionally, the policy trained through CoRS-DHC is compared with other advanced MAPF algorithms. The current SOTA algorithm, DCC\cite{9665278}, which is also based on reinforcement learning, is selected as the main comparison object, with the centralized MAPF algorithm ODrM*\cite{ferner2013odrm} and PRIMAL\cite{sartoretti2019primal} (based on RL and IL) serving as references. DCC is an efficient model designed to enhance agent performance by training agents to selectively communicate with their neighbors during both training and execution stages. It introduces a complex decision causal unit to each agent, which determines the appropriate neighbors for communication during these stages. Conversely, the PRIMAL algorithm achieves distributed MAPF by imitating ODrM* and incorporating reinforcement learning algorithms. ODrM* is a centralized algorithm designed to generate optimal paths for multiple agents. It is one of the best centralized MAPF algorithms currently available. We use it as a comparative baseline to show the differences between distributed and centralized algorithms.
The experimental results are demonstrated in Fig.~\ref{results}:
\begin{figure}[t]
\begin{minipage}{\linewidth}
    \centerline{\includegraphics[width=\linewidth]{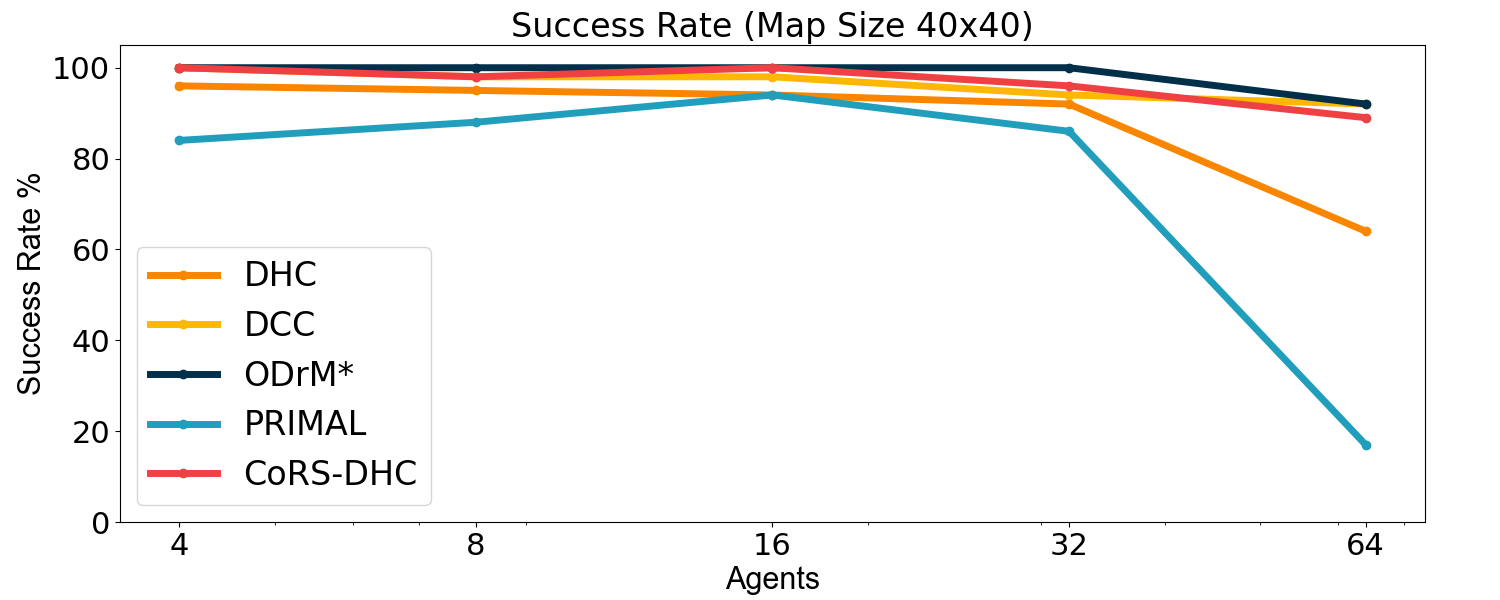}}
\end{minipage}
    
\begin{minipage}{\linewidth}
    \centerline{\includegraphics[width=\linewidth]{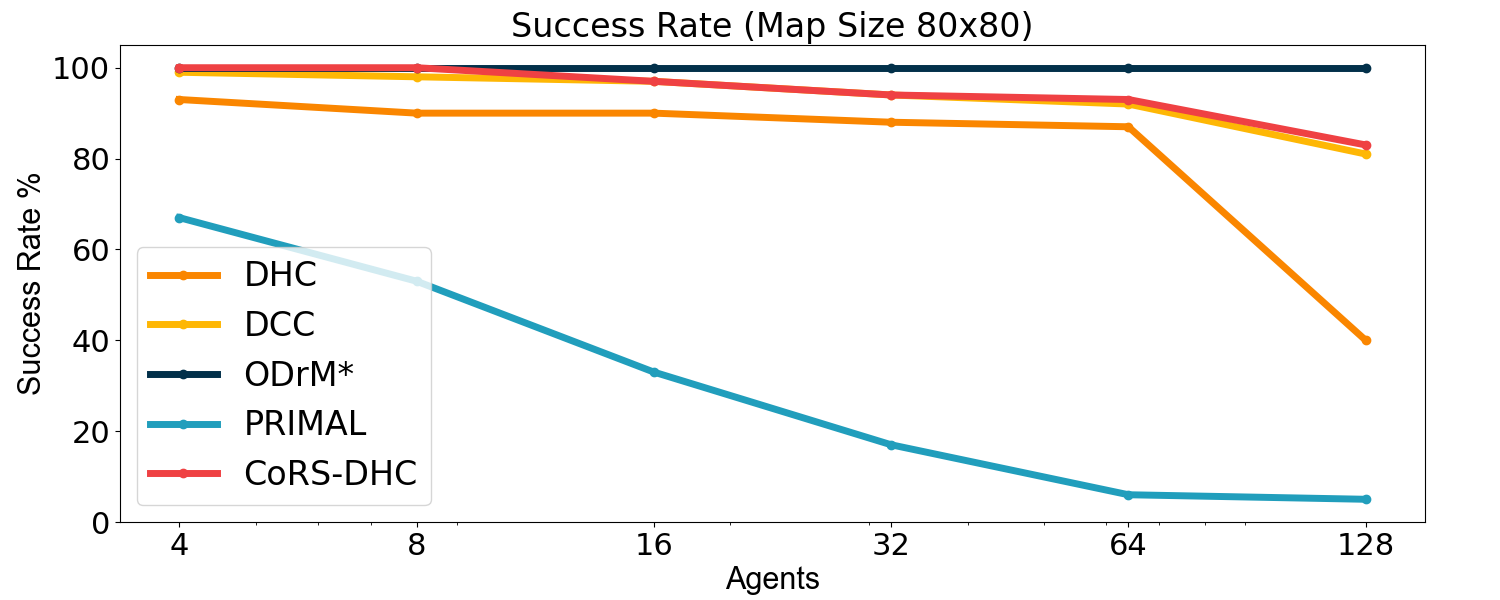}}
\end{minipage}

\begin{minipage}{\linewidth}
    \centerline{\includegraphics[width=\linewidth]{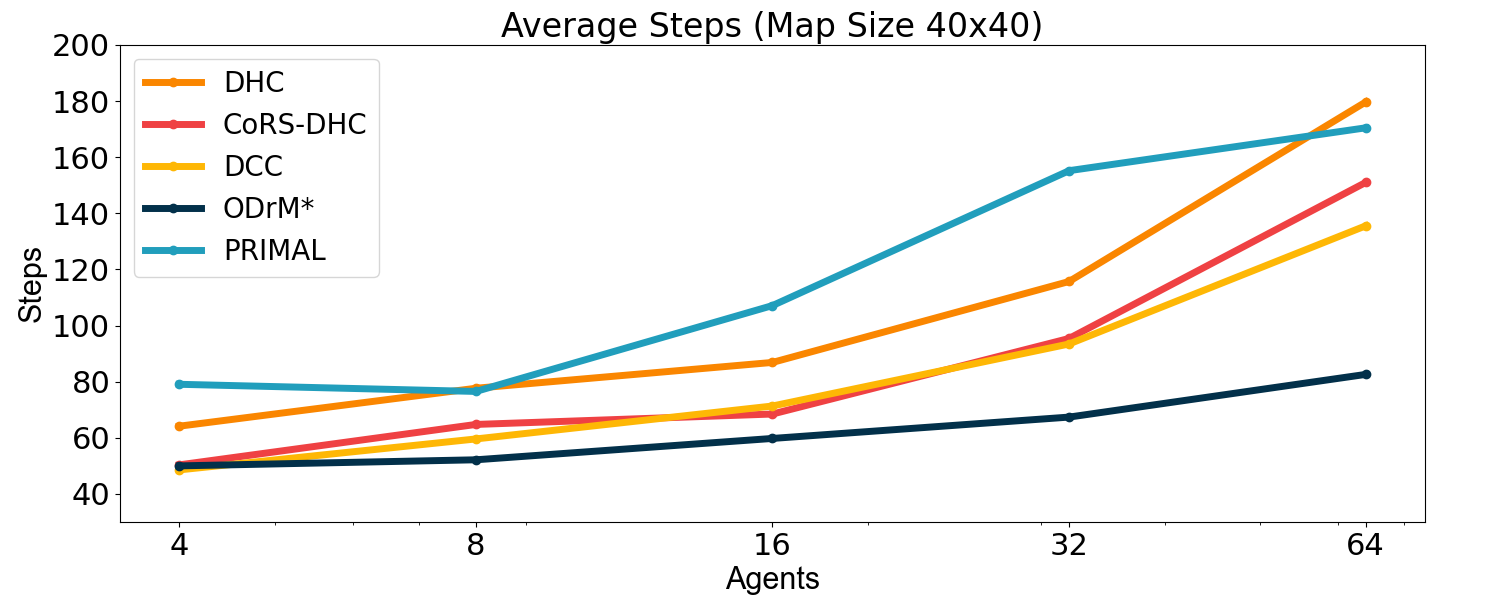}}
\end{minipage}
    
\begin{minipage}{\linewidth}
    \centerline{\includegraphics[width=\linewidth]{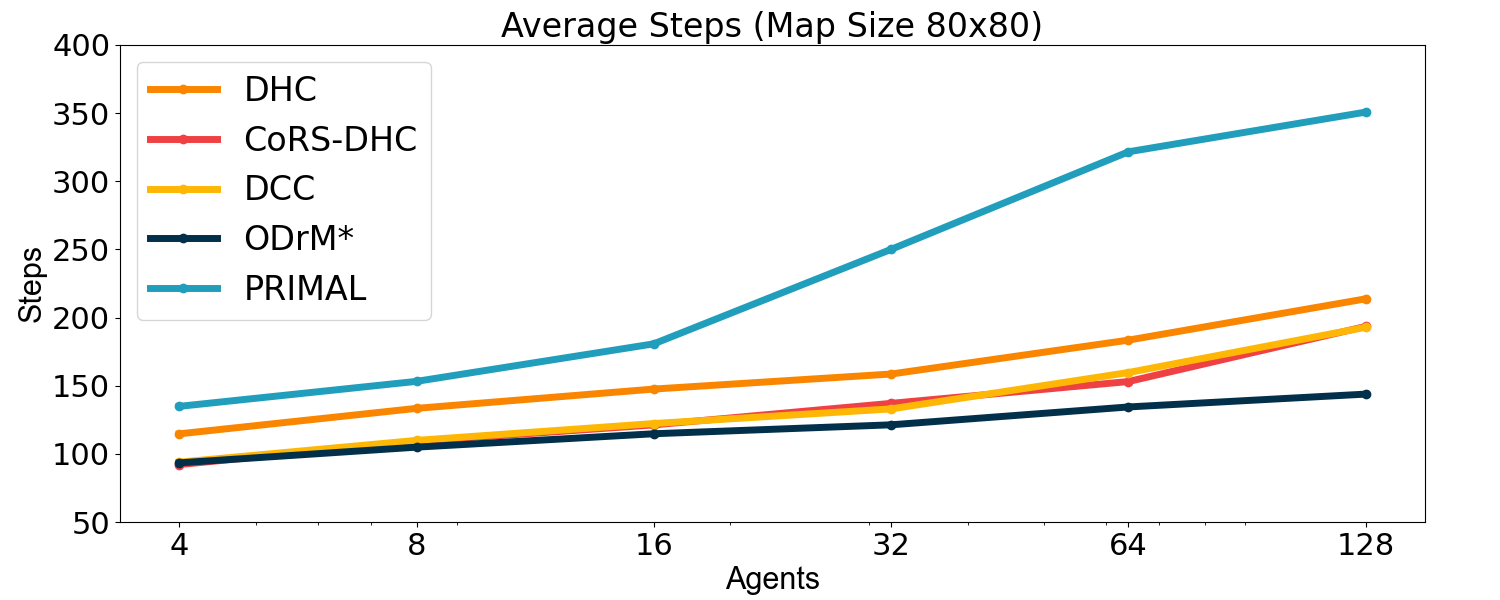}}
\end{minipage}
\vspace{-0.5cm}
\caption{Success rate and average steps across different testing scenarios.}
\label{results}
\end{figure}

The experimental results indicate that our CoRS-DHC algorithm consistently exceeds the success rate of the DCC algorithm in the majority of scenarios. Additionally, aside from the $40 \times 40$ grid with 64 agents, the makespan of the policies trained by the CoRS-DHC algorithm is comparable to or even shorter than that of the DCC algorithm across other scenarios. These results clearly demonstrate that our CoRS-DHC algorithm achieves a performance comparable to that of DCC. However, it should be noted that DCC employs a significantly more complex communication mechanism during both training and execution, while our CoRS algorithm only utilizes simple reward shaping during the training phase. Compared to PRIMAL and DHC, CoRS-DHC exhibits a remarkably superior performance.

\section{Conclusion} \label{Conclusion}
This letter proposes a reward shaping method termed \textbf{CoRS}, applicable to the standard MAPF tasks. By promoting cooperative behavior among multiple agents, CoRS significantly improves the efficiency of MAPF. The experimental results indicate that CoRS significantly enhances the performance of the MARL algorithms in solving the MAPF problem. CoRS also has implications for other multi-agent reinforcement learning tasks. We plan to further extend the application of this reward-shaping strategy to a wider range of MARL environments in future exploration.

\bibliographystyle{unsrt}
\bibliography{reference.bib}

\clearpage
\appendices

\setcounter{theorem}{0} 
\setcounter{definition}{0} 
\setcounter{lemma}{0} 
\setcounter{property}{0}
\setcounter{assumption}{0}

\section{Proofs} \label{proofs}
In this section, proofs of Theorem~\ref{Q IGM} is provided. For clarity, we hereby reiterate the assumptions involved in the proof processes.

\begin{assumption}
    The reward for agents staying at the target point is $0$, while the rewards for both movement and staying at non-target points are $r_m < 0$. The collision reward $r_c < r_m$.
\end{assumption}
\begin{assumption}
    During the interaction process between $A^i$ and $\tilde{A}^{-i}$, neither $A^i$ nor $\tilde{A}^{-i}$ has ever reached its respective endpoint.
\end{assumption}


\begin{theorem} \label{Q IGM}
    Assume Assumps.~\ref{assumption 1} and~\ref{assumption 2} hold. Then when $\alpha = \frac{1}{2}$, $\mathcal{Q}^i_{\pi^i_*} (\bar{s}_t, a^i)$, $\mathcal{Q}^{-i}_{\pi^{-i}_*} (\bar{s}_t, a^{-i})$ and $Q^{tot}_{\bar{\pi}_*} (\bar{s}_t, \{a^i, a^{-i}\})$ satisfy the \textit{IGM condition}.
\end{theorem}

\begin{proof}
    Theorem~\ref{Q IGM} is equivalent to the following statement: for $a^i_* = \operatorname{argmax}\limits_{a^i} \mathcal{Q}^i_{\pi^i_*} (\bar{s}_t, a^i)$, there exists $a^{-i}_* = \operatorname{argmax}\limits_{a^{-i}} \mathcal{Q}^{-i}_{\pi^{-i}_*} (\bar{s}_t, a^{-i})$, such that $Q^{tot}_{\bar{\pi}_*} (\bar{s}_t, \{a^i_*, a^{-i}_*\}) = \operatorname{argmax}\limits_{\bar{a}} Q^{tot}_{\bar{\pi}_*} (\bar{s}_t, \bar{a})$.
    
    For the sake of simplicity in the discussion, we denote $r^i(\bar{s}_t, \bar{a}_t)$ and $r^{-i}(\bar{s}_t, \bar{a}_t)$ as $r^i_t$ and $r^{-i}_t$, and $\mathbb{E}_{\tau} [\sum_{i=0}^{N} \gamma^t r^j_t]$ as $\sum_{\tau} \gamma^t r^j$. When $\alpha = \frac{1}{2}$, $\tilde{r}^i = \frac{1}{2}r^i + \frac{1}{2} \max_{a^{-i}} r^{-i}$ and $\tilde{r}^{-i} = \frac{1}{2}r^{-i} + \frac{1}{2} \max_{a^i} r^{i}$. 

    First, under Assumps.~\ref{assumption 1} and Assump.~\ref{assumption 2}, if no collisions occur along the trajectory $\tau$, then for all $\bar{s}_t$ and $\bar{a}_t \in \tau$, we have $r^i = \max_{a^i} r^i = r_m$ and $r^{-i} = \max_{a^{-i}} r^{-i} = r_m$. This holds because if $A^i$ and $\tilde{A}^{-i}$ do not reach their goals, then $r^i < 0$ and $r^{-i} < 0$. Meanwhile, if no collisions occur along $\tau$, then $r^i > r_c$ and $r^{-i} > r_c$. Thus, it follows that for all $\bar{s}_t$ and $\bar{a}_t \in \tau$, $r^i = \max_{a^i} r^i = r_m$ and $r^{-i} = \max_{a^{-i}} r^{-i} = r_m$.

    Second, $\forall a^i \in \mathcal{A}^i$ and $\forall \{a^i, a^{-i}\} \in \mathcal{A}$, $\mathcal{Q}^i_{\pi^i_*} (\bar{s}_t, a^i) \ge \frac{1}{2} Q^{tot}_{\bar{\pi}_*}(\bar{s}_t, \{a^i, a^{-i}\})$. This is because: 
    \begin{equation*}
        \begin{aligned}
            \mathcal{Q}_{\pi^i_*}^i& (\bar{s}, a^i) - \frac{1}{2} Q_{\bar{\pi}_*}^{tot}(\bar{s}, \bar{a}) \\
            =& \sum_{\tau^i} \gamma^t \left(  \frac{1}{2} r^i + \frac{1}{2} \max_{a^{-i}} r^{-i} \right) - \sum_{\bar{\tau}} \gamma^t \left( \frac{1}{2} r^i + \frac{1}{2} r^{-i} \right) \\
            \overset{(a)}{\ge}& \sum_{\bar{\tau}} \gamma^t \left(  \frac{1}{2} r^i + \frac{1}{2} \max_{a^{-i}} r^{-i} \right) - \sum_{\bar{\tau}} \gamma^t \left( \frac{1}{2}r^i + \frac{1}{2}r^{-i} \right)\\
            =& \sum_{\bar{\tau}} \gamma^t \left( \frac{1}{2} \max_{a^{-i}} r^{-i} - \frac{1}{2} r^{-i} \right) \ge 0 ,
        \end{aligned}
    \end{equation*} 
    where (a) holds because the trajectory $\tau^i$ maximizes $\sum \gamma^t \tilde{r}^i$, when the trajectory $\tau^i$ is replaced by $\bar{\tau}$, $\sum_{\tau^i} \gamma^t \tilde{r}^i \ge \sum_{\bar{\tau}} \gamma^t \tilde{r}^i$.

    Then consider $a^i_*$ and $\mathcal{Q}^i_{\pi^i_*} (\bar{s}, a^i_*) = \sum_{\tau^i} \gamma^t (\frac{1}{2} r^i + \frac{1}{2} \max_{a^{-i}} r^{-i})$. There must be no collisions along $\tau^i$, otherwise $a^i_*$ cannot be optimal. Therefore, $\max_{a^i} \mathcal{Q}^i_{\pi^i_*} (\bar{s}_t, a^i) = \sum_{\tau^i} \gamma^t (\frac{1}{2} r^i + \frac{1}{2} \max_{a^{-i}} r^{-i}) = \sum_{\tau^i} \gamma^t (\frac{1}{2} r^i + \frac{1}{2} r^{-i})$. Furthermore, since $\mathcal{Q}^i_{\pi^i_*} (\bar{s}_t, a^i_*) \ge \frac{1}{2} Q^{tot}_{\bar{\pi}_*}(\bar{s}_t, \{a^i_*, a^{-i}\})$, we find that $\sum_{\tau^i} \gamma^t (\frac{1}{2} r^i + \frac{1}{2} r^{-i}) = \max_{a^i} \mathcal{Q}^i_{\pi^i_*} (\bar{s}_t, a^i) = \frac{1}{2} \max_{\bar{a}} Q^{tot}_{\bar{\pi}_*} (\bar{s}_t, \bar{a}) = \frac{1}{2} Q^{tot}_{\bar{\pi}_*} (\bar{s}_t, \bar{a}_t)$, where $\bar{a}_t \in \tau^i$. We select $a^{-i}_*$ such that $\{a^i_*, a^{-i}_*\} = \bar{a}_t$. 

    Based on the previous discussion, it can be deduced that $a^i_*$ and $a^{-i}_*$ can maximize both $\mathcal{Q}^i_{\pi^i_*}(\bar{s}_t, a^i)$ and $Q^{tot}_{\bar{\pi}^i_*} (\bar{s}_t, \{a^i, a^{-i}\})$. Next, we demonstrate that $a^{-i}_*$ can also maximize $\mathcal{Q}^{-i}_{\pi^{-i}_*} (\bar{s}_t, a^{-i})$. 
    Let $\mathcal{Q}^i_{\pi^i_*}(\bar{s}_t, a^i_*) = Q^{tot}_{\bar{\pi}^i_*}(\bar{s}_t, \{a^i_*, a^{-i}_*\}) = U$. 
    Given that there is no conflict in $\tau^i$ and $U$ represents the maximum accumulated reward of $\frac{1}{2} r^i + \frac{1}{2} r^{-i}$ in state $\bar{s}t$, we have $U = \sum_{\tau^i} \gamma^t \left( \frac{1}{2} r^i + \frac{1}{2} r^{-i} \right) = \sum_{\tau^i} \gamma^t \left( \frac{1}{2} \max_{a^i} r^i + \frac{1}{2} r^{-i} \right) = \mathcal{Q}^{-i}_{\pi^{-i}_*} (\bar{s}, a^{-i}_*)$.
    Assume there exists $\hat{a}^{-i} \ne a^{-i}_*$ such that $\max_{a^{-i}}\mathcal{Q}^{-i}_{\pi^{-i}_*}(\bar{s}_t, a^{-i}) = \mathcal{Q}^{-i}_{\pi^{-i}_*}(\bar{s}_t, \hat{a}^{-i}) = V > U$. 
    Therefore:
    \begin{equation*}
        \begin{aligned}
            U &< V = \sum_{\tau^{-i}} \gamma^t (\frac{1}{2} r^{-i} + \frac{1}{2} \max_{a^i} r^i) = \sum_{\tau^{-i}} \gamma^t (\frac{1}{2} r^{-i} + \frac{1}{2} r^i) \\ 
            &\overset{(b)}{\le} \sum_{\tau^{i}} \gamma^t (\frac{1}{2} r^{-i} + \frac{1}{2} r^i) = U,
        \end{aligned}
    \end{equation*}
    which is clearly a contradiction. The validity of (b) follows from that $\sum_{\tau^i} \gamma^t \left( \frac{1}{2} r^i + \frac{1}{2} r^{-i} \right) = \frac{1}{2} \max_{\bar{a}} Q^{tot}_{\bar{\pi}_*} (\bar{s}_t, \bar{a})$. 
    Therefore, $\mathcal{Q}^{-i}_{\pi^{-i}_*}(\bar{s}_t, a^{-i}_*) = \max_{a^{-i}} \mathcal{Q}^{-i}_{\pi^{-i}_*}(\bar{s}_t, a^{-i})$.

    
\end{proof}

\section{Examples} \label{examples}
In this section, some examples will be provided to facilitate the understanding of the CoRS algorithm.

\subsection{The instability of Eq.~\eqref{original Ic}} \label{subsection:instability}
\begin{figure}[h]
    \includegraphics[width=\linewidth]{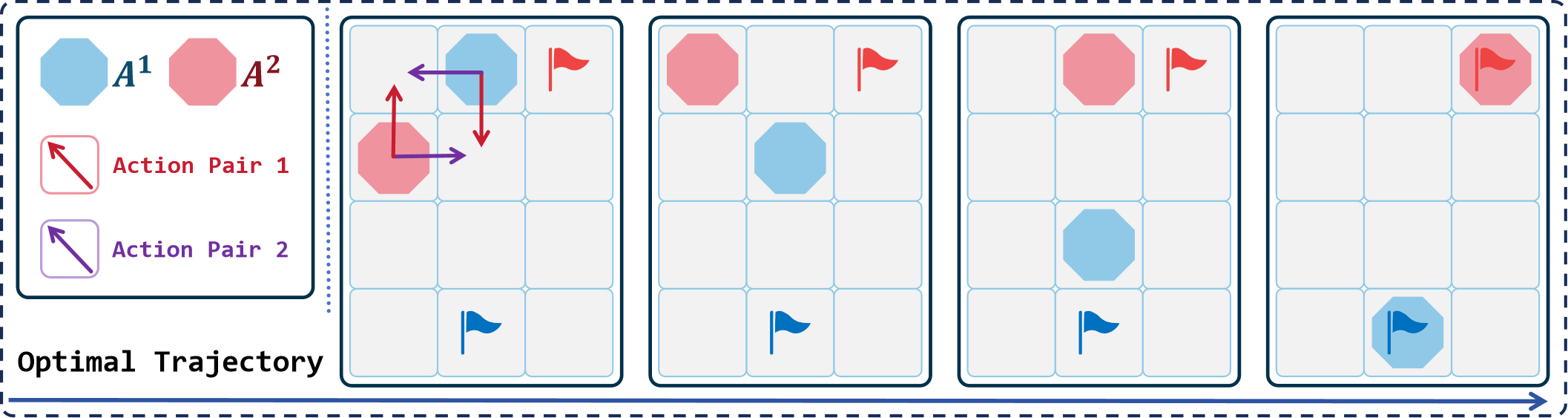}
    \vspace{-0.7cm}
    \caption{An example illustrating the instability of Eq.~\eqref{original Ic}.}
    \label{fig:instablity}
\end{figure}

In this section, we illustrate the instability of Eq.~\eqref{original Ic} through an example. Fig.~\ref{fig:instablity} presents two agents, $A^1$ and $A^2$, with the blue and red flags representing their respective endpoints. The optimal trajectory in this scenario is already depicted in the figure. Based on the optimal trajectory, it is evident that the optimal action pair in the current state is \textbf{Action Pair 1}. We anticipate that when $A^1$ takes the action from \textbf{Action Pair 1}, $I^1_c(\bar{s}_t, \bar{a}_t)$ will be higher than for other actions.

According to the calculation method provided in Eq.~\eqref{original Ic} and the reward function given in Table~\ref{reward function}, we compute the values of $I^1_c$ when $A^1$ takes the action from \textbf{Action Pair 1} under different conditions. When $A^2$ takes the action in \textbf{Action Pair 1}, $I^1_c = \frac{1}{|A^{-1}|} \sum_{j \in A^{-1}} r^j = r^2 = -0.070$. When $A^2$ takes the action in \textbf{Action Pair 2}, a collision occurs between $A^1$ and $A^2$, and $I^1_c = \frac{1}{|A^{-1}|} \sum_{j \in A^{-1}} r^j = r^2 = -0.5$. These calculations show that in the same state, even though $A^1$ takes the optimal action, the value of $I^1_c$ changes depending on $a^2$. This indicates that the calculation method provided in Eq.~\eqref{original Ic} is unstable.


\subsection{How Reward Shaping Promotes Inter-Agent Cooperation} \label{appendix:B-C}

\begin{figure*}[t]
    \centering
    \includegraphics[width=0.85\linewidth]{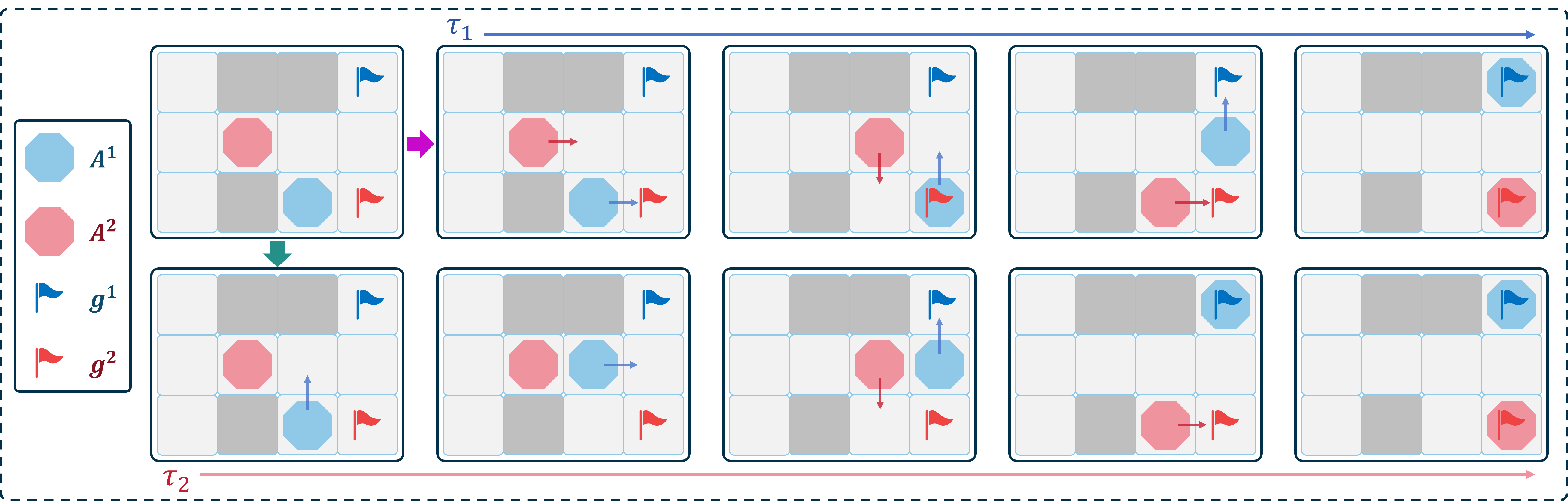}
    \vspace{-0.2cm}
    \caption{An example illustrating how reward shaping method enhances inter-agent cooperation.}
    \label{fig:cumulate}
\end{figure*}

In this section, we illustrate with an example how the reward-shaping method Eq.~\eqref{reward shaping method} promotes inter-agent cooperation, thereby enhancing the overall system efficiency. 
Table~\ref{reward function} presents the reward function used in this example. 
As shown in Fig.~\ref{fig:cumulate}, there are two agents, $A^1$ and $A^2$, engaged in interaction. Throughout their interaction, $A^1$ and $A^2$ remain neighbors. 
Each agent aims to maximize its cumulative reward. For this two-agent scenario, we set $\alpha = \frac{1}{2}$. Two trajectories, $\tau_1$ and $\tau_2$, are illustrated in Fig.~\ref{fig:cumulate}. We will calculate the cumulative rewards that $A^1$ can obtain on the two trajectories using different reward functions. For the sake of simplicity, in this example, $\gamma = 1$.

If $A^1$ uses $r^1$ as the reward function, the cumulative reward on the trajectory $P^1$ is $(-0.070) + (-0.070) + (-0.070) = -0.21$, and on the trajectory $P^2$ it is $(-0.070) + (-0.070) + (-0.070) = -0.21$. Consequently, $A^1$ may choose trajectory $\tau_2$. In $\tau_2$, $A^2$ will stop and wait due to being blocked by $A^1$.
This indicates that using the original reward function may not promote collaboration between agents, resulting in a decrease in overall system efficiency.

However, if $A^1$ adopts $\tilde{r}^1 = \frac{1}{2} r^1 + \frac{1}{2} \max_{a^2} r^2$ as its reward, the cumulative rewards for $A^1$ on trajectories $\tau_1$ is $(-\frac{1}{2}0.070 -\frac{1}{2} \cdot 0.070) + (-\frac{1}{2} \cdot 0.070 -\frac{1}{2} \cdot 0.070) + (-\frac{1}{2} \cdot 0.070 -\frac{1}{2} \cdot 0.070) = -0.21$. In contrast, on trajectory $\tau_2$, it is $(-\frac{1}{2} \cdot 0.070 -\frac{1}{2} \cdot 0.075) + (-\frac{1}{2} \cdot 0.070 -\frac{1}{2} \cdot 0.070) + (-\frac{1}{2} \cdot 0.070 -\frac{1}{2} \cdot 0.070) + (\frac{1}{2} \cdot 0 -\frac{1}{2} \cdot 0.070) = -0.2475$. 
Therefore, $A^1$ will choose the trajectory $\tau_1$ instead of $\tau_2$, allowing $A^2$ to reach its goal as quickly as possible. This demonstrates that using the reward $\tilde{r}^1$ shaped by the reward-shaping method Eq.~\eqref{reward shaping method} enables $A^1$ to adequately consider the impact of its actions on other agents, thereby promoting cooperation among agents and enhancing the overall efficiency of the system.

\section{Details of Experiments}
In this section, we provide detailed data in Section \ref{Experiments}. Table~\ref{table:success rate 40} and Table~\ref{table:success rate 80} presents the success rates of different algorithms across various test scenarios, while Table~\ref{table:makespan 40} and Table~\ref{table:makespan 80} shows the makespan of different algorithms in these scenarios. The results for the ODrM* algorithm, which are considered optimal, are highlighted in \textbf{bold}. The best metrics among other algorithms that use reinforcement learning techniques are marked in \red{red}.

\begin{table}[H]
    \centering
    \caption{Success Rate in 40 $\times$ 40 Environment.} \label{table:success rate 40}
    \begin{tabular}{c|c|c|c|c|c}
         \toprule
         \hline
         Success Rate\% & \multicolumn{5}{c}{Map Size $40 \times 40$} \\ 
         \hline
         Agents & DHC & DCC & CoRS-DHC              & PRIMAL & ODrM*          \\ \hline
         4      & 96  & 100 & \textcolor{red}{100}  & 84     & \textbf{100}   \\ 
         8      & 95  & 98  & \textcolor{red}{98}   & 88     & \textbf{100}   \\ 
         16     & 94  & 98  & \textcolor{red}{100}  & 94     & \textbf{100}   \\
         32     & 92  & 94  & \textcolor{red}{96}   & 86     & \textbf{100}   \\
         64     & 64  & \textcolor{red}{91}  & 89   & 17     & \textbf{92}    \\
         \hline
         \bottomrule
    \end{tabular}
\end{table}
\begin{table}[H]
    \centering
    \caption{Success Rate in 80 $\times$ 80 Environment.} \label{table:success rate 80}
    \begin{tabular}{c|c|c|c|c|c}
         \toprule
         \hline
         Success Rate\% & \multicolumn{5}{c}{Map Size $80 \times 80$} \\ 
         \hline
         Agents & DHC & DCC & CoRS-DHC             & PRIMAL & ODrM* \\ \hline
         4      & 93  & 99  & \textcolor{red}{100} & 67     & \textbf{100}   \\ 
         8      & 90  & 98  & \textcolor{red}{100} & 53     & \textbf{100}   \\ 
         16     & 90  & 97  & \textcolor{red}{97}  & 33     & \textbf{100}   \\
         32     & 88  & 94  & \textcolor{red}{94}  & 17     & \textbf{100}   \\
         64     & 87  & 92  & \textcolor{red}{93}  & 6      & \textbf{100}   \\
         128    & 40  & 81  & \textcolor{red}{83}  & 5      & \textbf{100}   \\
         \hline
         \bottomrule
    \end{tabular}
\end{table}

\begin{table}[H]
    \centering
    \caption{Average Steps in 40 $\times$ 40 Environment.} \label{table:makespan 40}
    \begin{tabular}{c|c|c|c|c|c}
         \toprule
         \hline
         Average Steps & \multicolumn{5}{c}{Map Size $40 \times 40$} \\ 
         \hline
         Agents & DHC    & DCC    & CoRS-DHC                 & PRIMAL  & ODrM*          \\ \hline
         4      & 64.15  & \textcolor{red}{48.58}  & 50.36   & 79.08   & \textbf{50}    \\ 
         8      & 77.67  & \textcolor{red}{59.60}  & 64.77   & 76.53   & \textbf{52.17} \\ 
         16     & 86.87  & 71.34  & \textcolor{red}{68.48}   & 107.14  & \textbf{59.78} \\
         32     & 115.72 & \textcolor{red}{93.45}  & 95.42   & 155.21  & \textbf{67.39} \\
         64     & 179.69 & \textcolor{red}{135.55} & 151.02  & 170.48  & \textbf{82.60} \\
         \hline
         \bottomrule
    \end{tabular}
\end{table}

\begin{table}[H]
    \centering
    \caption{Average Steps in 80 $\times$ 80 Environment.} \label{table:makespan 80}
    \begin{tabular}{c|c|c|c|c|c}
         \toprule
         \hline
         Average Steps & \multicolumn{5}{c}{Map Size $80 \times 80$} \\ 
         \hline
         Agents & DHC     & DCC     & CoRS-DHC                & PRIMAL & ODrM*             \\ \hline
         4      & 114.69  & 93.89   & \textcolor{red}{92.14}  & 134.86 & \textbf{93.40}    \\ 
         8      & 133.39  & 109.89  & \textcolor{red}{109.15} & 153.20 & \textbf{104.92}   \\ 
         16     & 147.55  & 122.24  & \textcolor{red}{121.25} & 180.74 & \textbf{114.75}   \\
         32     & 158.58  & \textcolor{red}{132.99} & 137.06  & 250.07 & \textbf{121.31}   \\
         64     & 183.44  & 159.67  & \textcolor{red}{153.06} & 321.63 & \textbf{134.42}   \\
         128    & 213.75  & \textcolor{red}{192.90} & 193.50 & 350.76 & \textbf{143.84}   \\
         \hline
         \bottomrule
    \end{tabular}
\end{table}

\end{document}